\documentclass[11pt]{article}
\usepackage{coling2016}
\usepackage{times}
\usepackage{url}
\usepackage{latexsym}
\usepackage{enumitem}
\usepackage{comment}

\title{Modeling Trolling in Social Media Conversations}

\author{Luis Gerardo Mojica\\
  Human Language Technology Research Institute\\
  The University of Texas at Dallas\\
  Richardson, TX 75083-0688\\
  mojica@utdallas.edu}

\date{}

\begin{document}
\maketitle
\begin{abstract}
Social media websites, electronic newspapers and Internet forums allow visitors to leave comments for others to read and interact. This exchange is not free from participants with malicious intentions, who {\em troll} others by positing messages that are intended to be provocative, offensive, or menacing.
With the goal of facilitating the computational modeling of trolling,
we propose a trolling categorization that 
is novel in the sense that it allows comment-based analysis from both the trolls' and the responders' perspectives, characterizing these two perspectives using four aspects, namely, the troll's intention and his intention disclosure, as well as the responder's interpretation of the troll's intention and her response strategy. Using this categorization, we annotate and release a dataset containing excerpts of Reddit conversations involving suspected trolls and their interactions with other users. Finally, we identify the difficult-to-classify cases in our corpus and suggest potential solutions for them.

\end{abstract}

\section{Introduction} 

In contrast to traditional content distribution channels like television, radio and newspapers, Internet opened the door for direct interaction between the content creator and its audience. 
Young people
are now gaining more frequent
access to online, networked media. Although
most of the time, their Internet use is
harmless, there are some risks associated with these
online activities, such as the use of social networking
sites (e.g., Twitter, Facebook, Reddit). The anonymity and freedom
provided by social networks makes them
vulnerable to threatening situations on the Web,
such as {\em trolling}.

Trolling is ``the activity of posting messages via a communications network that 
are intended to be provocative, offensive or menacing'' \cite{bishop2013effect}.
People who post such comments are known as {\em trolls}.
According to \newcite{hardaker2010trolling}, a troll's ``real intention(s) is/are to cause disruption and/or trigger or exacerbate conflict for the purpose of their own amusement''.
Worse still, the troll's comments may have a negative psychological impact on his target/victim and possibly others who participated in the same conversation. 
It is therefore imperative to identify such comments and perhaps even terminate the conversation before it evolves into something psychological disruptive for the participants. Monitoring conversations is a labor-intensive task: it can potentially place a severe burden on the moderators, and it may not be an effective solution when traffic is heavy. This calls for the need to develop automatic methods for identifying malicious comments, which we will refer to as {\em trolling attempts} in this paper.

In fact, there have recently been some attempts to automatically identify comments containing  {\em cyberbullying} (e.g., \newcite{van2015detection}), which corresponds to the most severe cases of trolling \cite{bishop2013effect}.
However, we believe that it is important not only to identify trolling attempts, but also comments that could have a negative psychological impact on their recipients. 
As an example, consider the situation where a commenter posts a comment with the goal of amusing others. However, it is conceivable that not everybody would be aware of these playful intentions, and these people may disagree or dislike the mocking comments and take them as inappropriate, prompting a negative reaction or psychological impact on themselves. 

In light of this discussion, we believe that there is a need to identify not only the trolling attempts, but also comments that could have a negative psychological impact on its receipts.
To this end, we seek to achieve the following goals in this paper.
First, we propose a comprehensive categorization of trolling that allows us to model not only the troll's intention given his trolling attempt, but also the recipients' {\em perception} of the troll's intention and subsequently their reaction to the trolling attempt.
This categorization gives rise to very interesting problems in pragmatics that involve the computational modeling of intentions, perceived intentions, and reactions to perceived intentions.
Second, we create a new annotated resource for computational modeling of trolling. Each instance in this resource corresponds to a {\em suspected} trolling attempt taken from a Reddit conversation, it's surrounding context, and its immediate responses and will be manually coded with information such as the troll's intention and the recipients' reactions using our proposed categorization of trolling.
Finally, we identify the instances that are difficult to classify with the help of a classifier trained with features taken from the state of the art, and subsequently present an analysis of these instances.

To our knowledge, our annotated resource is the first one of its sort that allows computational modeling on both the troll's side and the recipients' side. By making it publicly available, we hope to stimulate further research on this task. We believe that it will be valuable to any NLP researcher who is interested in the computational modeling of trolling.

\section{Related Work}

In this section, we discuss related work in the areas of trolling, bullying, abusive language detection and politeness, as they intersect in their scope and at least partially address the problem presented in this work.

In the realm of psychology, \newcite{bishop2013effect} and \newcite{bishop2014representations} elaborate a deep description of a troll's personality, motivations, effects on the community that trolls interfere in and the criminal and psychological aspects of trolls. Their main focus are flaming (trolls), and hostile and aggressive interactions between users \cite{o2003reconceptualizing}.

On the computational side, \newcite{mihaylov2015finding} address the problem of identifying manipulation trolls in news community forums. Not only do they focus solely on troll identification, but the major difference with this work is that all their predictions are based on non-linguistic information such as number of votes, dates, number of comments and so on. 
In a networks related framework, \newcite{kumar2014accurately} and \newcite{guha2004propagation} present a methodology to identify malicious individuals in a network based solely on the network's properties rather than on the textual content of comments.
\newcite{cambria2010not} propose a method that involves NLP components, but fail to provide an evaluation of their system. 

There is extensive work on detecting offensive and abusive language in social media \cite{nobata2016abusive} and \cite{xiang2012detecting}. There are two clear differences between their work and ours. One is that trolling is concerned about not only abusive language but also a much larger range of language styles and addresses the intentions and interpretations of the commenters, which goes beyond the linguistic dimension. The other is that we are additionally interested in the reactions to trolling attempts, real or perceived, because we argued that this is a phenomenon that occurs in pairs through the interaction of at least two individuals, which is different from abusive language detection. Also, \newcite{xu2012learning}, \newcite{xu2012fast} and \newcite{xu2013examination} address bullying traces. Bullying traces are self-reported events of individuals describing being part of bullying events, but we believe that the real impact of computational trolling research is not on analyzing retrospective incidents, but on analyzing real-time conversations. \newcite{chen2012detecting} use lexical and semantic features to determine sentence offensiveness levels to identify cyberbullying, offensive or abusive comments on Youtube. On Youtube as well, \newcite{dinakar2012common} identified sensitive topics for cyberbullying. \newcite{dadvar2014experts} used expert systems to classify between bullying and no bullying in posts. \newcite{van2015detection} predict fine-grained categories for cyberbullying, distinguishing between insults and threats and identified user roles in the exchanges. Finally, \newcite{hardaker2010trolling} argues that trolling cannot be studied using established politeness research categories.

\section{Trolling Categorization}

In this section, we describe our proposal of a comprehensive trolling categorization. While there have been attempts in the realm of psychology to provide a working definition of trolling (e.g., \newcite{hardaker2010trolling}, \newcite{bishop2014representations}), their focus is mostly on modeling the troll's behavior. For instance, \newcite{bishop2014representations} constructed a ``trolling magnitude'' scale focused on the severity of abuse and misuse of internet mediated communications. \newcite{bishop2013effect} also categorized trolls based on psychological characteristics focused on pathologies and possible criminal behaviors. In contrast, our trolling categorization seeks to model not only the troll's behavior but also the impact on the recipients, as described below.

Since one of our goals is to {\em identify} trolling events, our datasets will be composed of {\em suspected} trolling attempts (i.e., comments that are suspected to be trolling attempts). In other words, some of these suspected trolling attempts will be real trolling attempts, and some of them won't. So, if a suspected trolling attempt is in fact not a trolling attempt, then its author will {\em not} be a troll.


To cover both the troll and the recipients, we define a (suspected trolling attempt, responses) pair as the basic unit that we consider for the study of trolling, where ``responses'' are all the direct responses to the suspected trolling attempt. We characterize a (suspected trolling attempt, responses) pair using four aspects. Two aspects describe the trolling attempt: (1) \textbf{Intention \texttt{(I)}} (what is its author's purpose?), and (2) \textbf{Intention Disclosure \texttt{(D)}} (is its author trying to deceive its readers by hiding his real (i.e., malicious) intentions?). The remaining two aspects are defined on each of the (direct) responses to the trolling attempt: (1) \textbf{Intention Interpretation \texttt{(R)}} (what is the responder's perception of the troll's intention?), and (2) the \textbf{Response strategy \texttt{(B)}} (what is the responder's reaction?). Two points deserve mention. First, R can be different from I due to misunderstanding and the fact that the troll may be trying to hide his intention. Second, B is influenced by R, and the responder's comment can itself be a trolling attempt.
We believe that these four aspects constitute interesting, under-studied pragmatics tasks for NLP researchers.

The possible values of each aspect are described in Table \ref{tab:values}. As noted before, since these are {\em suspected} trolling attempts, if an attempt turns out not to be a trolling attempt, its author will not be a troll.

\begin{table*}[!ht]
\centering
\small
\begin{tabular}{|p{2.1cm}|p{12.1cm}| p{1.8cm}|}
\hline
{\bf Class} & {\bf Description} & {\bf Size \%} \\
\hline
\multicolumn{3}{l}{\textbf{Intention \texttt{(I)}}} \\
\hline
Trolling & The comment is malicious in nature, aims to disrupt, annoy, offend, harm or spread purposely false information & 53.6\% (537) \\
\hline
Mock Trolling or Playing & The comment is playful, joking, teasing or mocking others without the malicious intentions as in the Trolling class & 8.9\% (89) \\
\hline
No Trolling & A simple comment without malicious or playful intentions. & 37.7\% (375) \\
\hline
\multicolumn{3}{l}{\textbf{Intention Disclosure \texttt{(D)}}} \\
\hline
Exposed & A troll, clearly exposing its malicious or playful intentions & 34.7\% (347)\\
\hline
Hidden & A troll hiding its real malicious or playful intentions & 11.5\% (115)\\
\hline
None & The comment's author is not a troll, therefore there are no hidden nor exposed malicious or playful intentions & 53.8\% (539) \\
\hline 
\multicolumn{3}{l}{\textbf{Intentions Interpretation \texttt{(R)}}} \\
\hline
Trolling & The responder believes that the suspected troll is being malicious, annoying, offensive, harmful or attempts to spread false information & 59.7\% (785) \\
\hline
Mock Trolling or Playing & The responder believes that the suspected troll is being playful, joking, teasing or mocking without the malicious intentions & 5.3\% (70) \\
\hline
No Trolling & The responder believes that the suspected comment has no malicious intentions nor is playful, it is a simple comment. & 35.0\% (461) \\

\hline
\multicolumn{3}{l}{\textbf{Response Strategy \texttt{(B)}}} \\
\hline
Engage & Fall in the perceived provocation, showing an emotional response, upset or annoyed &26.9\% (354) \\
\hline
Praise & Acknowledge the perceived malicious or playful intentions and positively recognize the troll's ingenuity or ability & 3.0\% (39)\\
\hline
Troll & Acknowledge the perceived malicious and counter attack with a trolling attempt & 24.0\% (316) \\
\hline
Follow & Acknowledge the perceived malicious or playful intention and play along with the troll, further trolling & 3.0\% (39) \\
\hline
Frustrate & Acknowledges the perceived malicious or playful intentions and attempt to criticize or minimize them & 13.0\% (171)\\
\hline
Neutralize & Acknowledges the perceived malicious or playful intentions and give no importance to them & 9.5\% (125)\\
\hline
Normal & There is no perception or interpretation of a trolling attempt and the response is a standard comment & 20.7\% (272)\\
\hline
\end{tabular}
\caption{Classes for trolling aspects: Intention, Intention Disclosure, Intention Interpretation and Response Strategy. \textit{Size} refers to the percentage per class, in parenthesis is the total number of instances in the dataset.}\label{tab:values}
\vspace{-3mm}
\end{table*}

For a given (suspected trolling attempt, responses) pair, not all of the 189 (= $3 \times 3 \times 3 \times 7$) combinations of values of the four aspects are possible. There are logical constraints that limit plausible combinations: a) \textit{Trolling} or \textit{Playing Intentions} \texttt{(I)} must have \textit{Hidden} or \textit{Exposed} Intention Disclosure \texttt{(D)}, b) \textit{Normal} intentions \texttt{(I)} can only have \textit{None} Intention disclosure \texttt{(D)} and c) \textit{Trolling} or Playing interpretation \texttt{(R)} cannot have \textit{Normal} response strategy \texttt{(B)}.

\subsection{Conversation Excerpts}

To enable the reader to better understand this categorization, we present two example excerpts taken from the original (Reddit) conversations. The first comment on each excerpt, generated by author \texttt{C0}, is given as a minimal piece of context. The second comment, written by the author \texttt{C1} in italics, is the suspected trolling attempt. The rest of the comments comprise all direct responses to the suspected trolling comment. 

\noindent\textbf{Example 1.}
\vspace{-0.40mm}
\begin{itemize}[noitemsep,nolistsep]
  \small{\item[\texttt{C0:}] Yeah, cause that's what usually happens. Also, quit following me around, I don't want a boyfriend.}
  \begin{itemize}[noitemsep,nolistsep]
    \small{\item [\texttt{\textbf{C1}}:] \emph{I wasn't aware you were the same person.... I've replied to a number of stupid people recently, my bad}}
    \begin{itemize}[noitemsep,nolistsep]
      \small{\item[\texttt{C0:}] Trollname trollpost brotroll}
    \end{itemize}
  \end{itemize}
\end{itemize}
\vspace{-1.1mm}

In this example, \texttt{C1} is teasing of C0, expecting to provoke or irritate irritate, and he is clearly \emph{disclosing} her trolling intentions. In \texttt{C0}'s response, we see that he clearly believe that \texttt{C1} is trolling, since is directly calling him a ``brotroll'' and his \emph{response strategy} is frustrate the trolling attempt by denouncing \texttt{C1} troll's intentions ``trollpost'' and true identity ``brotroll''.

\noindent\textbf{Example 2.}
\vspace{-0.4mm}
\begin{itemize}[noitemsep,nolistsep]
  \small{\item[\texttt{C0:}]Please post a video of your dog doing this. The way I'm imagining this is adorable.}
  \begin{itemize}[noitemsep,nolistsep]
    \small{\item [\texttt{\textbf{C1}}:] I hope the dog gets run over by a truck on the way out of the childrens playground.
}
    \begin{itemize}[noitemsep,nolistsep]
      \small{\item[\texttt{C2:}] If you're going to troll, can you at least try to be a bit more 
        \small{\item[\texttt{C0:}] Haha I hope the cancer kills you.}
      convincing?}
    \end{itemize}
  \end{itemize}
\end{itemize}
\vspace{-1.3mm}

In this example, we observe that \texttt{C0}'s first comment is making a polite request (\emph{Please}). In return, \texttt{C1} answer is a mean spirited comment whose \emph{intention} is to disrupt and possible hurtful \texttt{C0}. Also, \texttt{C1}'s comment is not subtle at all, so his intention is clearly \emph{disclosed}. As for \texttt{C2}, she is clearly acknowledging \texttt{C1}'s trolling intention and her \emph{response strategy} is a criticism which we categorize as \emph{frustrate}. Now, in \texttt{C0}'s second comment, we observe that his \emph{interpretation} is clear, he believes that \texttt{C1} is \emph{trolling} and the negative effect is so tangible, that his response strategy is to \emph{troll} back or \emph{counter-troll} by replying with a comparable mean comment.



\section{Corpus and Annotation}

\textit{Reddit}\footnote{https://www.reddit.com/} is popular website that allows registered users (without identity verification) to participate in fora grouped by topic or interest. Participation consists of posting stories that can be seen by other users, voting stories and comments, and comments in the story's comment section, in the form of a forum. The forums are arranged in the form of a tree, allowing nested conversations, where the replies to a comment are its direct responses. We collected all comments in the stories' conversation in \textit{Reddit} that were posted in August 2015. Since it is infeasible to manually annotate all of the comments, we process this dataset with the goal of extracting threads that involve suspected trolling attempts and the direct responses to them. To do so, we used Lucene\footnote{https://lucene.apache.org/} to create an inverted index from the comments and queried it for comments containing the word ``\textbf{troll}'' with an edit distance of 1 in order to include close variations of this word, 
hypothesizing that such comments 
would be reasonable candidates of real trolling attempts. We did observe, however, that sometimes people use the word \emph{troll} to point out that another user is trolling. Other times, people use the term to express their frustration about a particular user, but there is no trolling attempt. Yet other times people simply discuss trolling and trolls without actually observing one. Nonetheless, we found that this search produced a dataset in which 44.3\% of the comments are real trolling attempts. Moreover, 
it is possible for commenters 
to believe that they are witnessing a trolling attempt and respond accordingly even where there is none due to misunderstanding. Therefore, the inclusion of comments that do not involve trolling would allow us to learn
what triggers a user's interpretation of trolling when it is not present and what kind of response strategies are used.

For each retrieved comment, we reconstructed the original conversation tree it appears in, from the original post (i.e., the root) to the leaves, so that its parent and children can be recovered\footnote{We removed the comments whose text had been deleted}. We consider a comment in our dataset a suspected trolling attempt if at least one of its immediate children contains the word \emph{troll}. For annotation purposes, we created snippets of conversations exactly like the ones shown in \textit{Example 1} and \textit{Example 2}, each of which consists of the parent of the suspected trolling attempt, the suspected trolling attempt, and all of the direct responses to the suspected trolling attempt. 


We had two human annotators who were trained on snippets (i.e., (suspected trolling attempt, responses) pairs) taken from 200 conversations and were allowed to discuss their findings. After this training stage, we asked them to independently label the four aspects for each snippet. 
We recognize that this limited amount of information is not always sufficient to recover the four aspects we are interested in, so we give the annotators the option to discard instances for which they couldn't determine the labels confidently. The final annotated dataset consists of 1000 conversations composed of 6833 sentences and 88047 tokens. The distribution over the classes per trolling aspect is shown in the table \ref{tab:results} in the column ``Size''.

Due to the subjective nature of the task we did not expect perfect agreement. However, on the 100 doubly-annotated snippets, we obtained substantial inter-annotator agreement according to Cohen's kappa statistic \cite{cohen1968weighted} for each of the four aspects: Intention: 0.788,  Intention Disclosure: 0.780, Interpretation: 0.797 and Response 0.776. In the end, the annotators discussed their discrepancies and managed to resolve all of them.

\section{Trolling Attempt Prediction}

In this section, we make predictions on the four aspects of our task, with the primary goal of identifying the errors our classifier makes (i.e., the hard-to-classify instances) and hence the directions for future work, and the secondary goal of estimating the state of the art on this new task using only shallow (i.e., lexical and wordlist-based) features.

\subsection{Feature Sets} \label{features}
For prediction we define two sets of features: (1) a basic feature set taken from Van Hee's~\shortcite{van2015detection} paper on cyberbullying prediction, and (2) an extended feature set that we designed using primarily information extracted from wordlists and dictionaries.

\subsubsection{Basic Feature Set} 

\noindent\textbf{N-gram features}. We encode each lemmatized and unlemmatized unigram and bigram collected from the training comments 
as a binary feature. In a similar manner, we include the unigram and bigram along with their POS tag as in \cite{xu2012learning}. 
To extract these features we used Stanford CoreNLP \cite{manning-EtAl:2014:P14-5}. 

\noindent\textbf{Sentiment Polarity}. The overall comment's emotion could be useful to identify the response and intention in a trolling attempt. So, we apply the Vader Sentiment Polarity Analyzer \cite{hutto2014vader} and include four features, one per each measurement given by the analyzer: positive, neutral, negative and a composite metric, each as a real number value.

\subsubsection{Extended Feature Set}

\noindent\textbf{Emoticons}. Reddit's comments make extensive use of emoticons. We argue that some emoticons are specifically used in trolling attempts to express a variety of emotions, which we hypothesize would be useful to identify a comment's intention, interpretation and response. For that reason, we use the emoticon dictionary developed \newcite{hogenboom2015exploiting}. We create a binary feature whose value is one if at least one of these emoticons is found in the comment.

\noindent\textbf{Harmful Vocabulary}. In their research on bullying, \newcite{nitta2013detecting} identified a small set of words that are highly offensive. 
We create a binary feature whose value is one if the comment contains at least one of these words.

\noindent\textbf{Emotions Synsets}. As in \newcite{xu2012fast}, we extracted all lemmas associated with each WordNet \cite{miller1995wordnet} \emph{synset} involving seven emotions (anger, embarrassment, empathy, fear, pride, relief and sadness) as well as the synonyms of these emotion words extracted from the English \newcite{merriam2004merriam} dictionary. We create a binary feature whose value is one if any of these synsets or synonyms appears in the comment.

\noindent\textbf{Swearing Vocabulary}. We manually collected 1061 swear words and short phrases from the internet, blogs, forums and smaller repositories \footnote{http://www.noswearing.com/}. The informal nature of this dictionary resembles the type of language used by flaming trolls and agitated responses, so we encode a binary feature whose value is one when at least one such swear word is found in the comment.

\noindent\textbf{Swearing Vocabulary in Username}. An interesting feature that is suggestive of the intention of a comment is the author's username. We found that abusive and annoying commenters contained cursing words in their usernames. So, we create a binary feature whose value is one if a swear word from the swearing vocabulary is found in their usernames.

\noindent\textbf{Framenet}. We apply the SEMAFOR parser \cite{das2014frame} to each sentence in every comment, and construct three different types of binary features: every frame name that is present in the sentence, the frame name and the target word associated with it, and the argument name along with the token or lexical unit in the sentence associated with it. We believe that some frames are especially interesting from the trolling perspective. We hypothesize that these features are useful for identifying trolling attempts in which semantic and not just syntactic information is required.

\noindent\textbf{Politeness cues}. \newcite{danescu2013computational} identified cues that signal polite and impolite interactions among groups of people collaborating online. Based on our observations of trolling examples, it is clear that flaming, hostile and aggressive interactions between users \cite{o2003reconceptualizing} and engaged or emotional responses would use impolite cues. In contrast, neutralizing and frustrating responses to the troll avoid falling in confrontation and their vocabulary tends to be more polite. So we create a binary feature whose value is one if at least one cue appears in the comment.

\noindent\textbf{GloVe Embeddings}. All the aforementioned features constitute a high dimensional bag of words (BOW). Word embeddings were created to overcome certain problems with the BOW representation, like sparsity, and weight in correlations of semantically similar words. For this reason, and following \newcite{nobata2016abusive}, we create a distributed representation of the comments by averaging the word vector of each lowercase token in the comment found in the Twitter corpus pre-trained GloVe vectors \cite{pennington2014glove}. The resulting comment vector representation is a 200 dimensional array that is concatenated with the existing BOW. 

\subsection{Results}

Using the features described in the previous subsection, we train four independent classifiers using logistic regression\footnote{We use the \emph{scikit-learn}\cite{scikit-learn} implementation}, one per each of the four prediction tasks. 
All the results are obtained using 5-fold cross-validation experiments. In each fold experiment, we use three folds for training, one fold for development, and one fold for testing. All learning parameters are set to their default values except for the regularization parameter, which we tuned on the development set.
In Table \ref{tab:results} the leftmost results column reports F1 score based on majority class prediction. The next section (\emph{Single Feature Group}) reports F1 scores obtained by using one feature group at a time. The goal of the later set of experiments is to gain insights about feature predictive effectiveness. The right side section (\emph{All features}) shows the system performance measured using recall, precision, and F-1 as shown when all features described in section \ref{features} are used.

The majority class prediction experiment is simplest baseline to which we can can compare the rest of the experiments. In order to illustrate the prediction power of each feature group independent from all others, we perform the ``Single Feature Group'', experiments. As we can observe in Table \ref{tab:results} there are groups of features that independently are not better than the majority baseline, for example, the emoticons, politeness cues and polarity are not better \emph{disclosure} predictors than the majority base. Also, we observe that only n-grams and GloVe features are the only group of features that contribute to more than a class type for the different tasks. Now, the ``All Features'' experiment shows how the interaction between feature sets perform than any of the other features groups in isolation. The accuracy metric for each trolling task is meant to provide an overall performance for all the classes within a particular task, and allow comparison between different experiments. In particular, we observe that GloVe vectors are the most powerful feature set, accuracy-wise, even better than the experiments with all features for all tasks except interpretation.

\begin{table*}[!th]
\centering
\begin{small}

\begin{tabular} {|c|p{4mm}|p{3mm}p{3mm}p{3mm}p{3mm}p{3mm}p{3mm}p{3mm}p{3mm}p{3mm}p{4.5mm}|p{3mm}p{3mm}p{4.5mm}|p{3.5mm}|} 
\hline
& \multicolumn{1}{c|}{} & \multicolumn{10}{c|}{Single Feature Group} & \multicolumn{3}{c|}{All Features} &\\
\hline
Aspect/Class & mjr & emt & hrm & syn  & swr & usr & frm & cue & pol & ngr & glv & R & P & F1 & Size \\ 
\hline
\multicolumn{11}{l}{\textbf{I: Intention}}\\
\hline
No trolling & 69.7 & 69.4 & 69.2 & 68.3 & 67.3 & 66.2 & 69.6 & 69.6 & 69.6 & 64.2 & 71.9 & 76.0 & 60.8 & 67.5 & 53\\
Trolling & 0.0 & 1.0 & 2.9 & 4.2 & 32.2 & 7.4 & 0.0 & 0.0 & 0.0 & 41.2 & 27.4 & 41.8 & 49.0 & 45.1 & 38\\
Playing & 0.0 & 0.0 & 0.0 & 0.0 & 0.0 & 0.0 & 0.0 & 0.0 & 0.0 & 0.0 & 0.0 & 0.0 & 0.0 & 0.0 & 9\\
\hline
Accuracy & 53.5 & 53.4 & 53.4 & 52.5 & 54.6 & 51.1 & 53.5 & 53.5 & 53.5 & 52.8 & 57.9 & - & - & 56.4& -\\
\hline
\multicolumn{11}{l}{\textbf{D: hidden}}\\
\hline
Hidden & 0.0 & 0.0 & 0.0 & 0.0 & 0.0 & 0.0 & 0.0 & 0.0 & 0.0 & 0.0 & 0.0 & 0.0 & 0.0 & 0.0 & 12\\
Exposed & 0.0 & 0.0 & 0.5 & 0.0 & 32.5 & 5.7 & 0.0 & 0.0 & 0.0 & 51.2 & 31.8 & 49.8 & 53.0 & 49.8 & 35\\
None & 69.9 & 69.3 & 69.5 & 69.6 & 70.3 & 69.2 & 69.7 & 69.7 & 69.7 & 66.5 & 72.0 & 76.3 & 61.5 & 76.3 & 53\\
\hline
Accuracy & 53.9 & 53.4 & 53.6 & 53.5 & 57.5 & 53.1 & 53.9 & 53.9 & 53.9 & 56.8 & 58.9 & - & - & 57.9 &-\\
\hline
\multicolumn{11}{l}{\textbf{R: Interpretation}}\\
\hline
No trolling & 0.0 & 4.0 & 0.5 & 1.0 & 4.7 & 0.0 & 0.0 & 0.0 & 0.0 & 50.8 & 38.7 & 63.0 & 56.3 & 63.0 & 34\\
Playing & 0.0 & 0.0 & 0.0 & 0.0 & 0.0 & 0.0 & 0.0 & 0.0 & 0.0 & 0.0 & 0.0 & 0.0 & 0.0 & 0.0 & 6\\
Trolling & 74.2 & 73.6 & 74.1 & 73.8 & 74.2 & 74.2 & 74.2 & 74.2 & 74.2 & 74.3 & 76.3 & 74.3 & 72.8 & 74.3 & 60\\
\hline
Accuracy & 59.2 & 58.4 & 59.2 & 58.5 & 59.2 & 59.2 & 59.2 & 59.2 & 59.2 & 64.4 & 64.9 & - & -  & 66.3 &-\\
\hline
\multicolumn{11}{l}{\textbf{R: Interpretation}}\\
\hline
Frustrate & 0.0 & 0.0 & 0.0 & 0.0 & 0.0 & 0.0 & 0.0 & 0.0 & 0.0 & 30.0 & 19.4 & 30.3 & 28.0 & 28.0 & 12\\
Troll & 0.0 & 2.7 & 0.9 & 4.5 & 28.4 & 0.0 & 0.0 & 0.0 & 0.0 & 33.6 & 27.3 & 36.3 & 34.5 & 34.5  &24\\
Follow & 0.0 & 0.0 & 0.0 & 0.0 & 0.0 & 0.0 & 0.0 & 0.0 & 0.0 & 18.8 & 6.1 & 66.8 & 14.5 & 14.5 & 3\\
Praise & 0.0 & 0.0 & 0.0 & 0.0 & 0.0 & 0.0 & 0.0 & 0.0 & 0.0 & 5.8 & 0.0 & 25.0 & 3.8 & 3.8 & 3\\
Neutralize & 0.0 & 0.0 & 0.0 & 0.0 & 0.0 & 0.0 & 0.0 & 0.0 & 0.0 & 37.0 & 26.8  & 38.3 & 43.8 & 43.8 & 9\\
Normal & 0.0 & 2.8 & 0.0 & 0.0 & 1.0 & 0.0 & 0.0 & 0.0 & 0.0 & 41.2 & 35.4 & 37.5 & 42.3 & 42.3 & 20\\
Engage & 43.8 & 44.5 & 43.9 & 43.2 & 42.7 & 43.7 & 43.7 & 43.7 & 43.7 & 40.5 & 49.9  & 42.0 & 45.5 & 45.5 & 29\\
\hline
Accuracy & 36.0 & 28.4 & 27.9 & 27.9 & 30.1 & 28.1 & 28.1 & 28.1 & 28.1 & 36.0 & 38.0 & - & -  & 37.5 & -\\
\hline
\multicolumn{11}{l}{\textbf{All Tasks Combined}}\\
\hline
Total Accuracy & 52.5 & 48.4 & 48.5 & 48.1 & 50.4 & 47.9 & 48.7 & 48.7 & 48.7 & 52.5 & 54.9 & - & - & 54.5& -\\
\hline
\end{tabular}
\end{small}
\caption{\emph{Experiments Results}. Below the ``mjr'' header, we report F1 scores the the majority class prediction we report F1 scores for the four aspects of trolling: Intention, Intentions Disclosure, Interpretation, and Response strategy. Also, below the ``Single Feature Group'' header, we report F1 scores as before, when the feature group indicated in the column headers is the only feature group used for classifier. The column headers abbreviations stand for: Emoticons, Harmful Vocabulary, Emotion Synsets, Swearing Vocabulary, Swearing Vocabulary in Usernames, Framenet, Politeness cues, n-grams (actual n-grams and n-grams appended with their corresponding part of speech tag) and Glove embeddings in that order. Below the ``All Features'' header we report Recall, Precision and F1 score, respectively, when all features are use for prediction. All experiments are performed using a logistic regression classifier per task. The last column reports the class distribution in percentage per task. The last row of each trolling aspect reports accuracy (the percentage of instances correctly classified). The last row in the table reports total accuracy, the percentage of correctly classified instances considering all aspects.} \label{tab:results}
\end{table*}

The overall Total Accuracy score reported in table \ref{tab:results} using the entire feature set is 549. This result is what makes this dataset interesting: there is still lots of room for research on this task. Again, the primary goal of this experiment is to help identify the difficult-to-classify instances for analysis in the next section.

\section{Error Analysis} 
\label{sub:error_analysis}

In order to provide directions for future work, we analyze the errors made by the classifier trained on the extended features on the four prediction tasks.\\
Errors on \textit{Intention} \texttt{(I)} prediction: 
The \textbf{lack of background} is a major problem when identifying trolling comments. For example, ``your comments fit well in Stormfront'' seems inoffensive on the surface. However, people who know that Stormfront is a white supremacist website will realize that the author of this comment had an annoying or malicious intention. But our system had no knowledge about it and simply predicted it as non-trolling. These kind of errors reduces recall on the prediction of trolling comments. A solution would be to include additional knowledge from anthologies along with a sentiment or polarity. One could modify NELL \cite{NELL-aaai15} to broaden the understanding of entities in the comments.\\
\textbf{Non-cursing aggressions and insults} This is a challenging problem, since the majority of abusive and insulting comments rely on profanity and swearing. The problem arises with subtler aggressions and insults that are equally or even more annoying, such as ``Troll? How cute.'' and ``settle down drama queen''. The classifier has a more difficult task of determining that these are indeed aggressions or insults.
This error also decreases the recall of trolling intention. A solution would be to exploit all the comments made by the suspected troll in the entire conversation 
in order to increase the chances of finding curse words or other cues that lead the classifier to correctly classify the comment as trolling. \\
Another source of error is the presence of \textbf{controversial topic words} such as ``black'',``feminism'', ``killing'', ``racism'', ``brown'', etc.\ that are commonly used by trolls.
The classifier seems too confident to classify a comment as trolling in the presence of these words, but in many cases they do not. 
In order to ameliorate this problem, one could create ad-hoc word embeddings by training glove or other type of distributed representation on a large corpus for the specific social media platform in consideration. From these vectors one could expect a better representation of controversial topics and their interactions with other words so they might help to reduce these errors.\\
Errors on \textit{Disclosure} \texttt{(D)} prediction: 
A major source of error that affects disclosure is the \textbf{shallow meaning representation} obtained from the BOW model even when augmented with the distributional features given by the glove vectors. For example, the suspected troll's comment ``how to deal with refugees? How about a bullet to the head'' is clearly mean-spirited and is an example of disclosed trolling. However, to reach that conclusion the reader need to infer the meaning of ``bullet to the head'' and that this action is desirable for a vulnerable group like migrants or refugees. This problem produces low recall for the disclosed prediction task. A solution for this problem may be the use of deeper semantics, where we represent the comments and sentences in their logical form and infer from them the intended meaning.\\
Errors on \textit{Interpretation} \texttt{(R)} prediction: it is a common practice from many users to \textbf{directly ask} the suspected troll if he/she is trolling or not. There are several variations of this question, such as ``Are you a troll?'' and ``not sure if trolling or not''. While the presence of a question like these seems to give us a hint of the responder's interpretation, we cannot be sure of his interpretation without also considering the context. One way to improve interpretation is to exploit the response strategy, but the response strategy in our model is predicted independently of interpretation. 
So one solution could be similar to the one proposed above for the disclosure task problem: jointly learning classifiers that predict both variables simultaneously. Another possibility is to use the temporal sequence of response comments and make use of older response interpretation as input features for later comments. This could be useful since commenters seem to influence each other as they read through the conversation.\\
Errors on \textit{Response Strategy} \texttt{(B)} prediction: In some cases there is a \textbf{blurry line between ``Frustrate'' and ``Neutralize''}. The key distinction between them is that there exists some \textit{criticism} in the \textit{Frustrate} responses towards the suspected troll's comment, while ``Neutralizing'' comments acknowledge that the suspected troll has trolling intentions, but gives no importance to them. For example, response comments such as ``oh, you are a troll'' and ``you are just a lame troll'' are examples of this subtle difference. The first is a case of ``neutralize'' while the second is indeed criticizing the suspected troll's comment and therefore a ``frustrate'' response strategy. This kind of error affects both precision and recall for these two classes.
A possible solution could be to train a specialized classifier to disambiguate 
between ``frustrate'' and ``neutralize'' only. \\
Another challenging problem is the \textbf{distinction between the classes ``Troll'' and ``Engage''}. This is true when the direct responder is intensely flared up with the suspected comment to the point that his own comment becomes a trolling attempt. A useful indicator for distinguishing these cases are the presence of insults, and to detect them we look for swear words, but as we noted before, there is no guarantee that swear words are used for insulting. This kind of error affects the precision and recall for the ``troll'' and ``engage'' classes. A solution to this problem may be the inclusion of longer parts of the conversation. It is typical in a troll-engaged comment scheme to observe longer than usual exchanges between two users, and the comments evolve in very agitated remarks. One may then use this information to disambiguate between the two classes.

\section{Conclusion and Future Work}

We presented a new view on the computational modeling of trolling in Internet fora where we proposed a comprehensive categorization of trolling attempts 
that for the first time
considers trolling from not only the troll's perspective but also
the responders' perspectives.
This categorization gives rise to four interesting pragmatics tasks that
involve modeling intensions, perceived intensions, and reactions.
Perhaps most importantly, we create an annotated dataset that we believe is the first of its sort. We intend to make publicly available with the hope of stimulating research on trolling.

\bibliographystyle{acl}
\bibliography{emnlp2016}

\end{document}